\documentclass[conference]{IEEEtran}

\usepackage[dvips]{graphicx}
\usepackage{stfloats}
\usepackage{algorithmic}
\usepackage{algorithm}

\begin{document}

\title{A New Heuristic for Feature Selection\\ by Consistent Biclustering}

\author{
\IEEEauthorblockN{A. Mucherino\IEEEauthorrefmark{1}
                  S. Cafieri\IEEEauthorrefmark{2}
}

\IEEEauthorblockA{}

\IEEEauthorblockA{\IEEEauthorrefmark{1}INRIA, Lille Nord Europe, France. \,
                 Email: {\tt\small antonio.mucherino@inria.fr}
}


\IEEEauthorblockA{\IEEEauthorrefmark{2}ENAC, Toulouse, France. \,
                  Email: {\tt\small sonia.cafieri@enac.fr}
}
}

\maketitle


\begin{abstract}
Given a set of data, biclustering aims at finding simultaneous partitions in biclusters of its samples and of the features which are used for representing the samples. Consistent biclusterings allow to obtain correct classifications of the samples from the known classification of the features, and vice versa, and they are very useful for performing supervised classifications. The problem of finding consistent biclusterings can be seen as a feature selection problem, where the features that are not relevant for classification purposes are removed from the set of data, while the total number of features is maximized in order to preserve information. This feature selection problem can be formulated as a linear fractional 0--1 optimization problem. We propose a reformulation of this problem as a bilevel optimization problem, and we present a heuristic algorithm for an efficient solution of the reformulated problem. Computational experiments show that the presented algorithm is able to find better solutions with respect to the ones obtained by employing previously presented heuristic algorithms.
\end{abstract}


\IEEEpeerreviewmaketitle

\section{Introduction}  \label{sec:intro}

Data mining techniques are nowadays much studied, because of the growing amount of data which is available and that needs to be analyzed. In particular, clustering techniques aim at finding suitable partitions of a set of samples in clusters, where data are grouped by following different criteria. The focus of this paper is biclustering, where samples and features in a given set of data are partitioned simultaneously.

Given a set of samples, each sample in the set can be represented by a sequence of features, which are supposed to be relevant for the samples. If a set of data contains $n$ samples which are represented by $m$ features, then the whole set can be represented by an $m \times n$ matrix $A$, where the samples are organized column by column, and the features are organized row by row. A {\it bicluster} is a submatrix of $A$, which can be equivalently defined as a pair of subsets $(S_r,F_r)$, where $S_r$ is a cluster of samples, and $F_r$ is a cluster of features. A {\it biclustering} is then a partition of $A$ in $k$ biclusters:
\begin{displaymath}
B = \{ (S_1,F_1), (S_2,F_2), \dots, (S_k,F_k) \} , 
\end{displaymath}
such that the following conditions are satisfied:
\begin{equation}  \label{equ:clS}
\bigcup_{r=1}^k S_r \equiv A, \qquad S_{\zeta} \cap S_{\xi} = \emptyset \quad 1 \le \zeta \ne \xi \le k ,
\end{equation}
\begin{equation}  \label{equ:clF}
\bigcup_{r=1}^k F_r \equiv A, \qquad F_{\zeta} \cap F_{\xi} = \emptyset \quad 1 \le \zeta \ne \xi \le k ,
\end{equation}
where $k \le \min(n,m)$ is the number of biclusters \cite{Busygin05,Hartigan}. Note that the conditions (\ref{equ:clS}) ensures that $B_S = \{S_1,S_2,\dots,S_k\}$ is a partition of the samples in disjoint clusters, while the conditions (\ref{equ:clF}) ensures that $B_F = \{F_1,F_2,\dots,F_k\}$ is a partition of the features in disjoint clusters.

We focus on the problem of finding biclusterings of the set of samples and of the set of features. When such biclusterings can be found, not only clusters of samples are obtained (as in standard clustering), but, in addition, the features causing the partition of samples in these clusters are also identified. This information is very interesting in many real-life applications. In particular, biclustering techniques are widely applied for analyzing gene expression data, where samples represent particular conditions (for example, the presence or absence of a disease), and each sample is represented by a sequence of gene expressions. In this case, finding out which features (genes) are related to the samples can help in discovering information about diseases \cite{Busygin09,Madeira}.

The concept of {\it consistent biclustering} is very important in this domain \cite{Busygin05}. Let us consider a set of samples, and let us suppose that a certain classification is assigned to such samples. In other words, we know a partition in clusters of these samples: $B_S = \{S_1,S_2,\dots,S_k\}$. A classification for the corresponding features, i.e. for the features used for representing these samples, can be obtained from $B_S$ (see Section~\ref{sec:biclust} for details). Let us refer to this partition of the features with  $B_F = \{F_1,F_2,\dots,F_k\}$. Then, the procedure can be inverted, and from the obtained classification $B_F$ of the features, another classification for the samples can be computed: $\hat{B}_S = \{\hat{S}_1,\hat{S}_2,\dots,\hat{S}_k\}$. In general, $B_S$ and $\hat{B}_S$ differ. In the event in which they instead coincide, the biclustering $B = \{ (S_1,F_1), (S_2,F_2), \dots, (S_k,F_k) \}$ is referred to as a consistent biclustering.

Consistent biclusterings can be used for classification purposes. Let us suppose that a training set is available for a certain classification problem. In other words, we suppose that a set of samples, whose classification is known, is available. From the classification of the samples, a classification of the features can be found, and then a certain biclustering, as explained above. If this biclustering is consistent, then the original classification of the samples in the training set can be reconstructed from the classification of its features. Therefore, the classification of these features can also be exploited for finding a classification for other samples, which originally have no known classification.

Unfortunately, sets of data allowing for consistent biclusterings are quite rare. There are usually features that are not relevant for the classification of the samples, which can easily bring to misclassifications. Because of experimental errors or noise, these features could be assigned to a bicluster or another, and this uncertainty causes errors in the classifications. For avoiding this, all the features that are not relevant must be removed. Therefore, we are interested in selecting a certain subset of features for which a consistent biclustering can be found. Since it is preferable to keep the loss of information as low as possible, the number of features to be selected has to be the maximum possible.

The feature selection problem related to consistent biclustering is NP-hard \cite{Kundakcioglu}. It can be formulated as a 0--1 linear fractional optimization problem, which can be very difficult to solve. In particular, for large (real-life) sets of data, the corresponding optimization problem is also large, and therefore there are no examples in the literature in which deterministic techniques have been employed. In \cite{Busygin05,Nahapatyan}, two heuristic algorithms have been proposed for solving the 0--1 linear fractional optimization problem arising in the context of feature selection by biclustering.

In this paper, we propose a new heuristic algorithm for solving this feature selection problem. We reformulate the optimization problem as a bilevel optimization problem, in which the inner problem is linear. Therefore, we use a deterministic algorithm for solving the inner problem, which is nested into a general framework where a heuristic strategy is employed. Our computational experiments show that the proposed heuristic algorithm is able to find subsets of features allowing for consistent biclusterings. The obtained results are compared to the ones reported in other publications \cite{Busygin05,Nahapatyan}: in general, the heuristic algorithm that we propose is able to find consistent biclusterings in which the number of selected features is larger.

The remaining of the paper is organized as follows. In Section~\ref{sec:biclust}, we develop the concept of consistent biclustering in more details, and we present the corresponding feature selection problem. In Section~\ref{sec:heuristic}, we reformulate this feature selection problem as a bilevel optimization problem and we introduce a heuristic algorithm for an efficient solution of the problem. Computational experiments on real-life sets of data are presented in Section~\ref{sec:experiments}, as well as a comparison to another heuristic algorithm. Conclusions are given in Section~\ref{sec:conclusions}.

\section{Consistent biclustering}  \label{sec:biclust}

Let $A$ be an $m \times n$ matrix related to a certain set of data, where samples are organized column by column, and their features are organized row by row. If a classification of the samples is known, then the {\it centroids} of each cluster, computed as the mean among all the members of the same cluster, can be computed. Let $C_S$ be the matrix containing all these centroids, organized column by column, where its generic element $c_{ir}^S$ refers to the $i^{th}$ feature of the centroid of the $r^{th}$ cluster of samples. Analogously, a matrix $C_F$ containing the centroids of the clusters related to a known classification of the features can be defined. The generic element $c_{jr}^F$ of the matrix $C_F$ refers to the $j^{th}$ sample related to the centroid of the $r^{th}$ cluster of features. Finally, the symbol $a_i$ refers to the $i^{th}$ row of the matrix $A$, i.e. to a feature, and the symbol $a^j$ refers to the $j^{th}$ column of $A$, i.e. to a sample. In the following discussion, $k$ represents the number of biclusters (known a priori), and $r \in \{ 1,2,\dots,k \}$ refers to the generic bicluster. The symbols $\hat{r}$ and $\xi$ are used for referring to biclusters having particular properties.

Let us suppose that a classification for the samples in $A$ is known. In other words, the following partition in $k$ clusters is available:
\begin{displaymath}
B_S = \{S_1,S_2,\dots,S_k\} .
\end{displaymath}
Starting from this classification, the matrix $C_S$ of centroids can be computed. Given a feature $a_i$, we can check the value of $c_{ir}^S$ for all the clusters. If, for a certain cluster $S_{\hat{r}}$, the element $c_{i\hat{r}}^S$ is the largest for any possible $r$, then $S_{\hat{r}}$ is the cluster in which the feature $a_i$ is mostly expressed. Therefore, it is reasonable to give to this feature the same classification as the samples in $S_{\hat{r}}$. Formally, it is imposed that:
\begin{equation}  \label{equ:biClustCond1}
a_i \in F_{\hat{r}} \quad \Longleftrightarrow \quad c_{i\hat{r}}^S > c_{i\xi}^S \quad
\forall \xi \in \{ 1,2,\dots,k \} \quad \xi \ne \hat{r} .
\end{equation}
Note that a complete classification of all the features can be obtained by imposing the equivalence (\ref{equ:biClustCond1}) for all $a_i$.

Let
\begin{displaymath}
B_F = \{F_1,F_2,\dots,F_k\}
\end{displaymath}
be the computed classification of the features. Starting from this classification, the matrix $C_F$ can be computed. In a similar way, a classification of the samples can be obtained by imposing the following equivalence:
\begin{equation}  \label{equ:biClustCond2}
a^j \in \hat{S}_{\hat{r}} \quad \Longleftrightarrow \quad c_{j\hat{r}}^F > c_{j\xi}^F \quad
\forall \xi \in \{ 1,2,\dots,k \} \quad \xi \ne \hat{r} .
\end{equation}

Let
\begin{displaymath}
\hat{B}_S = \{\hat{S}_1,\hat{S}_2,\dots,\hat{S}_k\}
\end{displaymath}
be the computed classification of the samples. In general, the two classifications $B_S$ and $\hat{B}_S$ are different from each other. If they coincide, then the partition in biclusters
\begin{displaymath}
B = \{ (S_1,F_1), (S_2,F_2), \dots, (S_k,F_k) \}
\end{displaymath}
is, by definition, a {\it consistent biclustering}. As already remarked in the Introduction, the classification of the features obtained from consistent biclusterings can be exploited for classifying samples with an unknown classification \cite{Busygin05}.

If a consistent biclustering exists for a certain set of data, then it is said to be {\it biclustering-admitting}. However, sets of data admitting consistent biclusterings are very rare. Therefore, features must be removed from the set of data for making it become biclustering-admitting \cite{Busygin05}. During this process, it is very important to remove the least possible number of features, in order to preserve the information in the set of data. In practice, a maximal subset of {\it good} features must be extracted from the initial set. The problem of finding the maximal consistent biclustering can be seen as a {\it feature selection} problem.

Let $f_{ir}$ be a binary parameter which indicates if the generic feature $a_i$ belongs to the generic cluster $F_r$ ($f_{ir} = 1$) or not ($f_{ir} = 0$). Let $x \equiv \{ x_1,x_2,\dots,x_m \}$ be a binary vector of variables, where $x_i$ is 1 if the feature $a_i$ is selected, and it is 0 otherwise. The problem of finding a consistent biclustering considering the maximum possible number of features can be formulated as follows:
\begin{equation}  \label{equ:biclustObjFun}
\max_x \left( f(x) = \sum_{i=1}^m x_i \right)
\end{equation}
subject, $\forall \hat{r},\xi \in \{ 1,2,\dots,k \}, \hat{r} \ne \xi, j \in S_{\hat{r}}$, to:
\begin{equation}  \label{equ:BiConstr}
\frac{\displaystyle\sum_{i=1}^m a_{ij} f_{i\hat{r}} x_i}{\displaystyle\sum_{i=1}^m f_{i\hat{r}} x_i} >
\frac{\displaystyle\sum_{i=1}^m a_{ij} f_{i\xi} x_i}{\displaystyle\sum_{i=1}^m f_{i\xi} x_i} .
\end{equation}
The generic constraint (\ref{equ:BiConstr}) ensures that the $\hat{r}$-th feature is the mostly expressed if it belongs to the cluster $(S_{\hat{r}},F_{\hat{r}})$. Note that the two fractions are used for computing the centroids of the clusters of features, and that the sums (at the numerators and at the denominators) only consider the selected features (each unselected feature is automatically discarded because $x_i = 0$). The reader is referred to \cite{Busygin05} for additional details.

In this context, other two optimization problems have also been introduced \cite{Nahapatyan}. They are extensions of the problem (\ref{equ:biclustObjFun})-(\ref{equ:BiConstr}), which have been proposed in order to overcome some problems related to data affected by noise. If a partition in clusters for the samples is available, then we can find a partition in clusters for the features. Each feature is therefore assigned to the cluster $F_{\hat{r}}$ if $c_{i\hat{r}}^S$ is the centroid with the largest value. Let us suppose that the following condition holds for a certain feature $a_i$:
\begin{displaymath}
\min_{\xi \ne \hat{r}} \{ c_{i\hat{r}}^S - c_{i\xi}^S \} \le \varepsilon
\end{displaymath}
where $\varepsilon$ is a small positive real number. If this is the case, small changes (i.e.: noise) in the data can bring to different partitions of the features, because the margin between $c_{i\hat{r}}^S$ and other centroids is very small.

In order to overcome this problem, the concepts of {\it $\alpha$-consistent biclustering} and {\it $\beta$-consistent biclustering} have been introduced in \cite{Nahapatyan}. They bring to the formulation of the following two optimization problems. The problem of finding an $\alpha$-consistent biclustering with a maximal number of features is equivalent to solving the optimization problem:
\begin{equation}  \label{equ:biclustObjFunAlpha}
\max_x \left( f(x) = \sum_{i=1}^m x_i \right)
\end{equation}
subject, $\forall \hat{r},\xi \in \{ 1,2,\dots,k \}, \hat{r} \ne \xi, j \in S_{\hat{r}}$, to:
\begin{equation}  \label{equ:BiConstrAlpha}
\frac{\displaystyle\sum_{i=1}^m a_{ij} f_{i\hat{r}} x_i}{\displaystyle\sum_{i=1}^m f_{i\hat{r}} x_i} > \alpha_j +
\frac{\displaystyle\sum_{i=1}^m a_{ij} f_{i\xi} x_i}{\displaystyle\sum_{i=1}^m f_{i\xi} x_i} ,
\end{equation}
where each $\alpha_j > 0$. Similarly, the problem of finding a $\beta$-consistent biclustering with a maximal number of features is equivalent to solving the optimization problem:
\begin{equation}  \label{equ:biclustObjFunBeta}
\max_x \left( f(x) = \sum_{i=1}^m x_i \right)
\end{equation}
subject, $\forall \hat{r},\xi \in \{ 1,2,\dots,k \}, \hat{r} \ne \xi, j \in S_{\hat{r}}$, to:
\begin{equation}  \label{equ:BiConstrBeta}
\frac{\displaystyle\sum_{i=1}^m a_{ij} f_{i\hat{r}} x_i}{\displaystyle\sum_{i=1}^m f_{i\hat{r}} x_i} > \beta_j \times
\frac{\displaystyle\sum_{i=1}^m a_{ij} f_{i\xi} x_i}{\displaystyle\sum_{i=1}^m f_{i\xi} x_i} ,
\end{equation}
where each $\beta_j > 1$. All the presented optimization problems are NP-hard \cite{Kundakcioglu}. The reader who is interested in more information on the formulation of these optimization problems can refer to \cite{Busygin05,Nahapatyan,Pardalos}. For a simple and ampler discussion on biclustering, refer to \cite{BOOK}.

The three optimization problems (\ref{equ:biclustObjFun})-(\ref{equ:BiConstr}), (\ref{equ:biclustObjFunAlpha})-(\ref{equ:BiConstrAlpha}) and (\ref{equ:biclustObjFunBeta})-(\ref{equ:BiConstrBeta}) are linear fractional 0--1 optimization problems. In \cite{Busygin05}, a possible linearization of the problem has been studied. However, the authors noted that currently available solvers for mixed integer programming are not able to solve the considered linearization, due to the large number of variables which are usually involved when dealing with real-life data. Therefore, they presented a heuristic algorithm for the solution of these problems, which is based on the solution of a sequence of linear 0--1 (non-fractional) optimization problems. Successively, in \cite{Nahapatyan}, another heuristic algorithm has been proposed, where a sequence of continuous linear optimization problems needs to be solved. The heuristic algorithm we propose is able to provide better solutions with respect to the ones provided by these two.

\section{An improved heuristic}  \label{sec:heuristic}

In the following discussion, only the optimization problem (\ref{equ:biclustObjFun})-(\ref{equ:BiConstr}) will be considered, because similar observations can be made for the other two problems. The computational experiments reported in Section~\ref{sec:experiments}, however, will be related to all three optimization problems.

We propose a reformulation of the problem (\ref{equ:biclustObjFun})-(\ref{equ:BiConstr}) as a bilevel optimization problem. To this aim, we substitute the denominators in the constraints (\ref{equ:BiConstr}) with new variables $y_r$, $r = 1,2,\dots,k$, where each $y_r$ is related to the generic bicluster. Then, we can rewrite the constraints (\ref{equ:BiConstr}) as follows:
\begin{equation}  \label{equ:BiConstrNew}
\frac{1}{y_{\hat{r}}} \sum_{i=1}^m a_{ij} f_{i\hat{r}} x_i > \frac{1}{y_\xi} \sum_{i=1}^m a_{ij} f_{i\xi} x_i .
\end{equation}
The constraints (\ref{equ:BiConstrNew}) must be satisfied for all $\hat{r},\xi \in \{ 1,2,\dots,k \}, \hat{r} \ne \xi$ and for all $j \in S_{\hat{r}}$.

Let us consider a set of values $\bar{y}_r$ of $y_r$, and also another proportional set of values $\breve{y}_r = \delta \bar{y}_r$, with $\delta > 0$. It is easy to see that, given certain values for the variables $x_i$, with $i = 1,2,\dots,m$, the constraints (\ref{equ:BiConstrNew}) are satisfied with $\bar{y}_r$ if and only if they are satisfied with $\breve{y}_r$. As an example, if $k=3$ and there is a consistent biclustering in which 20, 30 and 50 features are selected in the $k$ biclusters, then the constraints (\ref{equ:BiConstrNew}) are also satisfied if 0.20, 0.30 and 0.50, respectively, replace the actual number of features (in this example, the proportional factor $\delta$ is 0.01). For this reason, the variables $y_r$ can be used for representing the {\it proportions} among the cardinalities of the clusters of features. In the previous example, 20\% of the selected features are in the first bicluster, 30\% of the features in the second one, and 50\% in the last one. The variables $y_r$ can be bound in the real interval $[0,1]$, and the following constraint can be included in the optimization problem:
\begin{equation}  \label{equ:newConstr}
\sum_{r=1}^k y_r = 1 .
\end{equation} 

We introduce the function:
\begin{displaymath}
c(x,y_{\hat{r}},y_{\xi}) = \sum_{j \in S_{\hat{r}}}
\mid \frac{1}{y_\xi} \sum_{i=1}^m a_{ij} f_{i\xi} x_i - \frac{1}{y_{\hat{r}}} \sum_{i=1}^m a_{ij} f_{i\hat{r}} x_i \mid_+ ,
\end{displaymath}
where the symbol $|\cdot|_+$ represents the function which returns its argument if it is positive, and it returns 0 otherwise. As a consequence, the value of this function is positive if and only if the corresponding constraints (\ref{equ:BiConstrNew}) are not satisfied. Finally, we reformulate the optimization problem (\ref{equ:biclustObjFun})-(\ref{equ:BiConstr}) as the bilevel optimization problem:
\begin{equation}  \label{equ:bilevel1}
\min_{y} \left( g(x,y) = \sum_{\hat{r}=1}^k \sum_{\xi \ne \hat{r}}  c(x,y_{\hat{r}},y_{\xi}) \right)
\end{equation}
subject to:
\begin{equation}  \label{equ:bilevel2}
\begin{array}{rcl}
x & = & \arg \displaystyle\max_{x} \left( f(x) = \displaystyle\sum_{i=1}^m x_i \right) \\
{}  &   & {\rm subject~to~constraint~(\ref{equ:BiConstrNew})} , \\
\multicolumn{3}{l}{\displaystyle\sum_{r=1}^k y_r = 1 .} \\
\end{array}
\end{equation}

The objective function $g$ of the outer problem is the sum of several terms which correspond to the function $c(x,y_{\hat{r}},y_{\xi})$ for each $\hat{r}$ and $\xi \in \{1,2,\dots,k \}$, with $\xi \ne \hat{r}$. The minimization of all the terms of $g$ brings to the identification of biclusterings in which the constraints (\ref{equ:BiConstrNew}) are all satisfied. If this is the case, the found biclustering is consistent.

Algorithm~\ref{algbiclust} is a sketch of our heuristic algorithm for feature selection by consistent biclustering.
\begin{algorithm}[t]
\begin{algorithmic}
\item let $iter = 0$;
\item let $x_i = 1$, $\forall i \in \{1,2,\dots,m \}$;
\item let $y_r = \sum_i f_{ir} / m$, $\forall r \in \{1,2,\dots,k\}$;
\item let $range = starting\_range$;
\WHILE{($g(x,y) > 0$ and $range \le max\_range$)}
   \item let $iter = iter + 1$;
   \item solve the inner optimization problem (linear \& cont.);
   \IF{($g(x,y) > 0$)}
      \item increase $range$;
      \IF{($g(x,y)$ has improved)}
         \item $range = starting\_range$;
      \ENDIF
      \item let $r'$ = random in $\{1,2,\dots,k\}$;
      \item choose randomly $y_{r'}$ in $[y_{r'}-range,y_{r'}+range]$;
      \item let $r''$ = random in $\{1,2,\dots,k\}$ such that $r' \ne r''$;
      \item set $y_{r''}$ so that $\sum_r y_r = 1$;
   \ENDIF
\ENDWHILE
\end{algorithmic}
\caption{A heuristic algorithm for feature selection.}  \label{algbiclust}
\end{algorithm}
At the beginning, the variables $x_i$ are all set to 1, and the variables $y_r$ are set so that they represent the distribution of all the $m$ features among the $k$ clusters. Therefore, if the biclustering is already consistent, then the function $g$ is 0 with this choice for the variables, and all the features can be selected. In this case, the condition in the while loop is not satisfied and the algorithm ends.

At each step of the algorithm, the inner optimization problem is solved. It is a linear 0--1 optimization problem, and we consider its continuous relaxation, i.e. we allow the variables $x$ to take any real value in the interval $[0,1]$. Therefore, after a solution has been obtained, we substitute the fractional values of $x_i$ with 0 if $x_i \le 1/2$, or with 1 if $x_i > 1/2$. Moreover, in the experiments, the strict inequality of the constraints (\ref{equ:BiConstrNew}) is relaxed, so that the domains defined by the constraints are closed domains. In these hypotheses, the optimization problem can be solved by commonly used solvers for mixed integer linear programming (MILP). In our experiments, we employ the ILOG CPLEX solver (version 11) \cite{ILOG}.

After the solution of the inner problem, the function $g$ is evaluated. If the obtained values for the variables $x_i$, together with the used values for the variables $y_r$, correspond to a value for $g$ equal to 0, then the outer problem is also solved and the algorithm stops. Otherwise, some parameters and variables are modified in order to get ready for the next iteration of the algorithm.

The heuristic part of this algorithm takes inspiration from the Variable Neighborhood Search (VNS) \cite{Hansen,Mladenovic}, which is one of the most successful meta-heuristic searches for global optimization \cite{Talbi}. The variables $y_r$ are randomly modified during the algorithm: at each step, two of such variables $y_{r'}$ and $y_{r''}$ are chosen randomly so that $r' \ne r''$. Then, $y_{r'}$ is perturbed, and its value is chosen randomly in the interval centered in the previous value of $y_{r'}$ and with length $2 \times range$. As in VNS, the considered interval is relatively small during the first iterations, in order to focus the search in neighbors of the current variable values. Then, the interval is increased and increased. However, it is set back to its starting size when better solutions are found. By employing this strategy borrowed from VNS, every time there is a new improvement on the objective function value, the search is initially focused in neighbors of the current solution, and then it is extended to the whole search domain. When the considered interval gets too large ($max\_range$), then the search is stopped, because there are low probabilities to find better solutions. After having chosen a value for $y_{r'}$, a new value for $y_{r''}$ is computed so that the constraint on all the variables $y_r$ is satisfied. Note that, for values of $range$ large enough, the randomly computed $y_{r'}$ could be such that
\begin{displaymath}
\sum_{\forall r \ne r''} y_r > 1 .
\end{displaymath}
In this case, there are no possible values for $y_{r''}$ in $[0,1]$ for which the constraint (\ref{equ:newConstr}) can be satisfied. In order to overcome this issue, too large values for $range$ are avoided.

For its nature, the proposed heuristic algorithm can provide different solutions if it is executed more than once (with different seeds for the generator of random numbers). Therefore, the algorithm can be executed a given number of times and the best obtained solution can be taken into consideration. 

\section{Computational experiments}  \label{sec:experiments}

We implemented the presented heuristic algorithm for feature selection in AMPL \cite{AMPL}, from which the ILOG CPLEX11 solver is invoked for the solution of the inner optimization problem. Experiments are carried out on an Intel Core 2 CPU 6400 @ 2.13 GHz with 4GB RAM, running Linux.  

The first set of data that we consider is a set of gene expressions related to human tissues from healthy and sick (affected by cancer) patients \cite{Notterman}. This set of data is available on the web site of the Princeton University (see the paper for the web link). It contains 36 samples classified as {\it normal} or {\it cancer}, and each sample is specified through 7457 features. We applied our heuristic algorithm for finding a consistent biclustering for the samples and the features contained in this set of data. 

Table~\ref{tab:carcinoma} shows some computational experiments.
\begin{table}
\begin{scriptsize}
\begin{center}
\begin{tabular}{||c||c||c|c|c|c|c|c||}
\hline
$\alpha$ & 0 & 1 & 2 & 5 & 10 \\
\hline
$f(x)$ & 7450 & 7448 & 7444 & 7413 & 7261 \\
\hline
\end{tabular}
\end{center}
\begin{center}
\begin{tabular}{||c||c||c|c|c|c|c|c||}
\hline
$\beta$ & 1 & 1.01 & 1.50 & 2.00 & 3.00 \\
\hline
$f(x)$ & 7450 & 7450 & 7107 & 6267 & 5365 \\
\hline
\end{tabular}
\end{center}
\caption{Computational experiments on a set of samples from normal and cancer tissues. The features are selected by finding an $\alpha$-consistent or $\beta$-consistent biclustering.} 
\label{tab:carcinoma}
\end{scriptsize}
\end{table}
We found $\alpha$-consistent biclusterings and $\beta$-consistent biclusterings, with different values for $\alpha$ or $\beta$. Note that, even though for each sample a different $\alpha_j$ or $\beta_j$ can be considered, we use one unique value for $\alpha$ and $\beta$ in each experiment. In the table, the number of selected features $f(x)$ is given in correspondence with each experiment. 

When $\alpha = 0$ or $\beta = 1$ (consistent biclustering), after 4 iterations only (41 seconds of CPU time), our heuristic algorithm is able to provide the list of selected features, and thus to identify the (few) features to be removed in order to have a consistent biclustering. In particular, 7 features on 7457 need to be removed (and therefore 7450 features are selected).

The bilevel optimization problem to be solved gets harder in the case of $\alpha$-consistent and $\beta$-consistent biclustering. As expected, less features are selected when larger $\alpha$ or $\beta$ values are chosen, because the constraints (\ref{equ:BiConstrNew}) are more difficult to be satisfied. However, using larger values for $\alpha$ and $\beta$ allows for identifying the features that are actually important for the classification of the samples. The computational cost of our heuristic algorithm increases when larger $\alpha$ or $\beta$ values are used: some of the presented experiments need some minutes of CPU time to be performed.

The second real-life set of data we consider consists of samples from patients diagnosed with acute lymphoblastic leukemia (ALL) or acute myeloid leukemia (AML) diseases \cite{Golub} (to download the set of data, follow the link given in the reference). This set of data is divided in a training set, which we use for finding consistent biclusterings, and a validation set, which can be used for checking the quality of the classifications performed by using the features previously selected. The training set contains 38 samples: 27 ALL samples and 11 AML samples. The validation set contains 34 samples: 20 ALL samples and 14 AML samples. The total number of features in both sets of data is 7129. Since, in this case, a validation set is also available, we are able to validate the quality of the obtained biclusterings in correspondence with different values for the chosen parameter $\alpha$ or $\beta$.

The results of our experiments are in Table~\ref{tab:ALLAML}.
\begin{table}
\begin{scriptsize}
\begin{center}
\begin{tabular}{||c||c|c||c|c||}
\hline
{} & \multicolumn{2}{|c||}{Alg.~\ref{algbiclust}} & \multicolumn{2}{|c||}{Alg. in \cite{Nahapatyan}} \\
\hline\hline
$\alpha$ & $f(x)$ & $err$ & $f(x)$ & $err$ \\
\hline
 0 & 7081 & 2 & 7024 & 2 \\
10 & 7076 & 2 & 7024 & 2 \\
20 & 7075 & 2 & 7018 & 2 \\
30 & 7072 & 2 & 7014 & 2 \\
40 & 7068 & 2 & 7010 & 1 \\
50 & 7061 & 1 & 6959 & 1 \\
60 & 7046 & 1 & 6989 & 1 \\
70 & 6954 & 1 & 6960 & 1 \\
\hline\hline
$\beta$ & $f(x)$ & $err$ & $f(x)$ & $err$ \\
\hline
1.00 & 7081 & 2 & 7024 & 2 \\
1.05 & 7075 & 2 & 7017 & 2 \\
1.10 & 7068 & 2 & 7010 & 1 \\
1.20 & 7020 & 1 & 6937 & 1 \\
1.50 & 6590 & 1 & 6508 & 1 \\
2.00 & 5987 & 1 & 5905 & 1 \\
3.00 & 5527 & 2 & 5458 & 1 \\
5.00 & 5238 & 2 & 5173 & 2 \\
\hline
\end{tabular}
\end{center}
\caption{Computational experiments on a set of samples from patients diagnosed with ALL or AML diseases. The features are selected by finding an $\alpha$-consistent or $\beta$-consistent biclustering.} 
\label{tab:ALLAML}
\end{scriptsize}
\end{table}
The total number of features that are selected in each experiment is reported, together with the number  $err$ of misclassifications that occur when the samples of the validation set are classified accordingly with the classification of the features in the $\alpha$-consistent or $\beta$-consistent biclusterings. When $\alpha = 0$ or $\beta = 1$, our heuristic algorithm is able to find a consistent biclustering, but the selected features are not able to provide a correct classification for all the samples of the validation set ($err = 2$). This is due to the fact that the used data are probably noisy, because they have been obtained from an experimental technique. However, the number $err$ of misclassifications decreases when $\alpha$ or $\beta$ increase. For example, for $\alpha \ge 50$, there is only one misclassification for the samples of the validation set. 

In Table~\ref{tab:ALLAML}, we also compare the obtained results to the ones reported in \cite{Nahapatyan}. Our heuristic algorithm is able to provide better-quality solutions in the majority of the cases. In particular, for given choices of $\alpha$ or $\beta$, our heuristic algorithm is able to find biclusterings in which the total number of selected features is larger, except for only one experiment ($\alpha = 70$). These biclusterings allow to perform good-quality classifications ($err =$ 1 or 2), while a larger number of features in the set of data are preserved.

\section{Conclusions}  \label{sec:conclusions}

We proposed a reformulation for the linear fractional 0--1 optimization problem for feature selection by consistent biclustering. Our reformulation transforms the problem into a bilevel optimization problem, in which the inner problem is linear. We presented a heuristic algorithm for the solution of the reformulated problem, where the continuous relaxation of the inner problem is solved exactly at each iteration of the algorithm. Computational experiments showed that the proposed algorithm can solve feature selection problems by finding consistent, $\alpha$-consistent and $\beta$-consistent biclusterings of a given set of data. The results also showed that this algorithm is able to find better solutions with respect to the ones obtained by previously proposed heuristic algorithms. Future works will be devoted to suitable strategies for improving the efficiency of the proposed algorithm.

\vfill\eject

\end{document}